\documentclass[conference]{IEEEtran}
\IEEEoverridecommandlockouts

\usepackage{cite}
\usepackage{amsmath,amssymb,amsfonts}
\usepackage{algorithmic}
\usepackage{graphicx}
\usepackage{textcomp}
\usepackage{xcolor}
\usepackage{booktabs}
\def\BibTeX{{\rm B\kern-.05em{\sc i\kern-.025em b}\kern-.08em
    T\kern-.1667em\lower.7ex\hbox{E}\kern-.125emX}}
\begin{document}

\title{Nationwide EHR-Based Chronic Rhinosinusitis Prediction Using Demographic-Stratified Models}

\author{
\IEEEauthorblockN{\small Sicong Chang$^{\dagger}$ \thanks{$\dagger$ These authors contributed equally to this work.}}
\IEEEauthorblockA{\scriptsize
\textit{Electrical and Computer Engineering}\\
\textit{University of Houston}\\
Houston, USA\\
schang19@uh.edu
}
\and
\IEEEauthorblockN{\small Yidan Shen$^{\dagger}$}
\IEEEauthorblockA{\scriptsize
\textit{Electrical and Computer Engineering}\\
\textit{University of Houston}\\
Houston, USA\\
yshen20@uh.edu
}
\and
\IEEEauthorblockN{\small Justina Varghese}
\IEEEauthorblockA{\scriptsize
\textit{Otolaryngology}\\
\textit{Houston Methodist Hospital}\\
Houston, USA\\
jrvarghese@houstonmethodist.org
}
\and
\IEEEauthorblockN{\small Akshay R Prabhakar}
\IEEEauthorblockA{\scriptsize
\textit{Otolaryngology}\\
\textit{Houston Methodist Hospital}\\
Houston, USA\\
aprabhakar@houstonmethodist.org
}
\and
\IEEEauthorblockN{\small Sebastian Guadarrama-Sistos-Vazquez}
\IEEEauthorblockA{\scriptsize
\textit{Otolaryngology}\\
\textit{Houston Methodist Hospital}\\
Houston, USA\\
sguadarrama@houstonmethodist.org
}
\and
\IEEEauthorblockN{\small Jiefu Chen}
\IEEEauthorblockA{\scriptsize
\textit{Electrical and Computer Engineering}\\
\textit{University of Houston}\\
Houston, USA\\
jchen82@central.uh.edu
}
\and
\IEEEauthorblockN{\small Masayoshi Takashima}
\IEEEauthorblockA{\scriptsize
\textit{Otolaryngology}\\
\textit{Houston Methodist Hospital}\\
Houston, USA\\
mtakashima@houstonmethodist.org
}
\and
\IEEEauthorblockN{\small Omar G. Ahmed}
\IEEEauthorblockA{\scriptsize
\textit{Otolaryngology}\\
\textit{Houston Methodist Hospital}\\
Houston, USA\\
ogahmed@houstonmethodist.org
}
\and
\IEEEauthorblockN{\small Renjie Hu}
\IEEEauthorblockA{\scriptsize
\textit{Information Science Technology}, \textit{University of Houston} \\
\& \textit{Otolaryngology}, \textit{Houston Methodist Hospital}\\
Houston, USA\\
rhu7@uh.edu
}
\and
\IEEEauthorblockN{\small Xin Fu$^{*}$ \thanks{$*$ Corresponding author: Xin Fu}}
\IEEEauthorblockA{\scriptsize
\textit{Electrical and Computer Engineering}\\
\textit{University of Houston}\\
Houston, USA\\
xfu8@central.uh.edu
}
}
\maketitle

\begin{abstract}
Chronic rhinosinusitis (CRS) is a common heterogeneous inflammatory disorder that causes substantial morbidity and healthcare costs.
CRS is difficult to identify early from routine encounters, as symptom presentations overlap with common conditions such as allergic rhinitis, and heterogeneous phenotypes further obscure risk patterns. 
Prior predictive studies often rely on single-institutional cohorts , which reduce population-level generalizability. 
To overcome this, we leveraged nationwide longitudinal EHR data from the \textit{All of Us} Research Program to predict CRS diagnosis using two years of pre-diagnostic history. To address extreme feature sparsity and dimensionality in coded EHR data, we implemented a hybrid feature-selection pipeline that combines prevalence-based statistical screening with model-based importance ranking, compressing approximately 110,000 candidate codes into 100 interpretable features. To capture demographic heterogeneity, we trained demographic stratified models across six adult sex and life-stage subgroups with subgroup-specific hyperparameter tuning. 
Our framework achieved an overall AUC of 0.8461, improving discrimination by 0.0168 over the best baseline. These results demonstrate that routinely collected EHR data may support population-representative CRS risk stratification and inform earlier triage and referral prioritization in primary care.
\end{abstract}

\begin{IEEEkeywords}
Chronic rhinosinusitis, Electronic Health Records, Machine learning, Feature selection, Disease Screening.
\end{IEEEkeywords}

\section{Introduction}
Chronic rhinosinusitis (CRS) is a heterogeneous chronic inflammatory condition of the paranasal sinuses and nasal passage \cite{Fokkens2020EPOS, Rosenfeld2015AdultSinusitis}. 
CRS is a major source of chronic morbidity, affecting approximately 11–12\% of the population in the United States \cite{cdc_nchs_faststats_sinus_2025}, imposing a substantial disease burden, being associated with significant reductions in health-related quality of life and a national financial burden of \$22 billion (2014 USD) annually in the United States \cite{Smith2015Cost,Rudmik2017Economics}. 
Formal clinical diagnosis of CRS requires objective inflammatory evidence on nasal endoscopy or computed tomography (CT), as symptom-based assessment overestimates prevalence due to overlap with other inflammatory airway conditions such as allergic rhinitis \cite{deLoos2019Prevalence}. 
However, not all patients have access to imaging particularly in many primary care settings. Leveraging routinely collected electronic health record (EHR) data to identify potential CRS cases before specialist evaluation may help shift decision-making from subjective impression to evidence-driven prediction.


A recent study applied logistic regression using nasal-polyp PRS and achieved near-chance discrimination, suggesting that simple linear models without clinical or environmental covariates have limited utility for predicting CRS \cite{jiang2023risk}. In contrast, some studies utilize imaging such as CT and MRI to obtain a better result \cite{deLoos2019Prevalence, chen2023dlradiomics}; however, CT/MRI is not always available and limits generalizability.
Related work has also leveraged EHR features within established CRS cohorts to predict downstream outcomes. Ramakrishnan et al. compared multiple machine-learning models to classify olfactory dysfunction among CRS patients \cite{ramakrishnan2021olfactory}, and Nuutinen et al. estimated revision sinus surgery risk using EHR-derived features with interpretability analyses \cite{nuutinen2022revision}. Hopkins et al. validated SNOT-22 as a patient-reported outcome measure with a clinically interpretable minimal important difference \cite{hopkins2009psychometric}. However, they do not predict whether an individual will be diagnosed with CRS. Moreover, many are derived from single-institution or care-intense cohorts that can over-represent severe or refractory phenotypes and under-represent the broader heterogeneous population.

To address these challenges, we utilized a nationwide longitudinal EHR dataset from the \textit{All of Us} Research Program to predict CRS diagnosis using two years of prior medical history of participants. 
But coded clinical features are extremely high-dimensional, sparse, and dominated by weakly informative variables. This setting is often ill-posed relative to sample size, introducing noise and computational burdens that degrade predictive performance and increase the risk of overfitting. To mitigate these issues, we implemented a hybrid feature-selection pipeline that combines statistical prevalence-based screening with model-based feature importance selection, yielding a compact and interpretable feature set and reducing more than 110,000 candidate features to 100 key variables.

We observed that phenotypic patterns vary by age and sex in our analyses: where males showed an age-dependent shift toward predominantly structural and polypoid pathology, whereas females exhibited profiles enriched for mucosal inflammation and atopy, with age modulating the progression from acute to more systemic manifestations. And these observations are supported by prior CRS studies\cite{stevens2019dysregulation, lu2021sex, wu2018estrogen, anderson2022aging, hwang2019chronic}.
Motivated by these, we applied a demographic-specific modeling framework. Specifically, individuals were stratified into six subgroups defined by the intersection of sex and age-based adult life stages. Independently tailored predictive models were trained for each subgroup to effectively capture well-established differences in chronic inflammatory disease presentation and healthcare utilization.
Overall, this demographic-specific approach achieved an AUC of 0.8461 and improved discrimination by 0.0168 relative to the best baseline, demonstrating that explicitly modeling demographic heterogeneity yields measurable gains in CRS risk prediction.
The contributions of this paper are as follows:
\begin{itemize}
    \item We observe and leverage the nationwide diversity and richness of factors in the \textit{All of Us} EHR cohort.
    \item We introduce a hybrid feature selection framework that compresses more than 110,000 clinical codes to 100 key features.
    \item We propose demographic stratified modeling motivated by pronounced age-dependent sexual dimorphism in disease phenotypes and progression.
    \item We build six independent models to capture different risk factors in each subgroup.
    \item Our approach demonstrates improved discrimination both overall and across subgroups.
    \item Our approach may facilitate earlier triage and referral prioritization in clinical settings, particularly when diagnostic imaging is unavailable.
\end{itemize}

\section{Related Work}

Machine learning (ML) has been increasingly utilized to enhance the diagnosis, phenotyping, and prognostic assessment of CRS. Broadly, prior research focuses on two primary domains: approaches leveraging specialized imaging or genetic modalities, and those utilizing electronic health records (EHR).

\subsection{Genetic and Imaging-Based Approaches}

A targeted line of research focuses specifically on predicting the presence of CRS Within this scope, genetic susceptibility has been explored using polygenic risk scores (PRS) derived from large-scale biobank cohorts to estimate disease liability across diverse ancestries \cite{jiang2023risk}. Complementary approaches leverage objective imaging data from head CT and MRI to define CRS based on Lund-Mackay thresholds and quantify inflammatory burden beyond symptom reporting \cite{deLoos2019Prevalence}. Methodologically, linear frameworks such as logistic regression are typically employed in genetic studies to evaluate the predictive utility of cumulative risk alleles. In contrast, imaging-centric efforts frequently adopt deep learning and radiomics pipelines to select high-dimensional signatures from scans for precise phenotype classification \cite{chen2023dlradiomics}. However, a primary limitation of the works is the reliance on resource-intensive data modalities; obtaining genetic sequencing or imaging entails substantial costs and logistical burdens, rendering these approaches unfeasible for routine clinical practice. Furthermore, existing modeling frameworks exhibit algorithmic shortcomings: linear models often lack the complexity to capture the nonlinear interactions inherent in heterogeneous disease pathways, while deep learning suffers from limited interpretability.

\subsection{EHR Data Modeling}


A parallel body of work utilizes widely accessible EHR and routine clinical data for predictive modeling. For instance, multicenter prospective cohorts have been leveraged to distinguish olfactory dysfunction from normal smell in CRS patients \cite{ramakrishnan2021predicting}. Similarly, analysis of longitudinal EHR data has been used to predict the individual risk of revision sinus surgery \cite{nuutinen2022using}, demonstrating that incorporating longer historical observation windows significantly improves model performance. In these settings, nonlinear learners such as random forest and decision trees are frequently adopted to capture higher-order interactions among symptoms and comorbid factors, while remaining relatively interpretable.

However, these studies rely on relatively small datasets, which presents two significant limitations. First, because they focus exclusively on characterizing outcomes within an established CRS cohort, these models cannot address the broader challenge of disease screening in the general population. Second, the restricted scale and scope of these cohorts introduce selection bias and lack population-level coverage, making it difficult to cross-validate findings or ensure generalizability across diverse demographic groups.

\begin{figure}[t]
    \centering
    \includegraphics[width=\columnwidth]{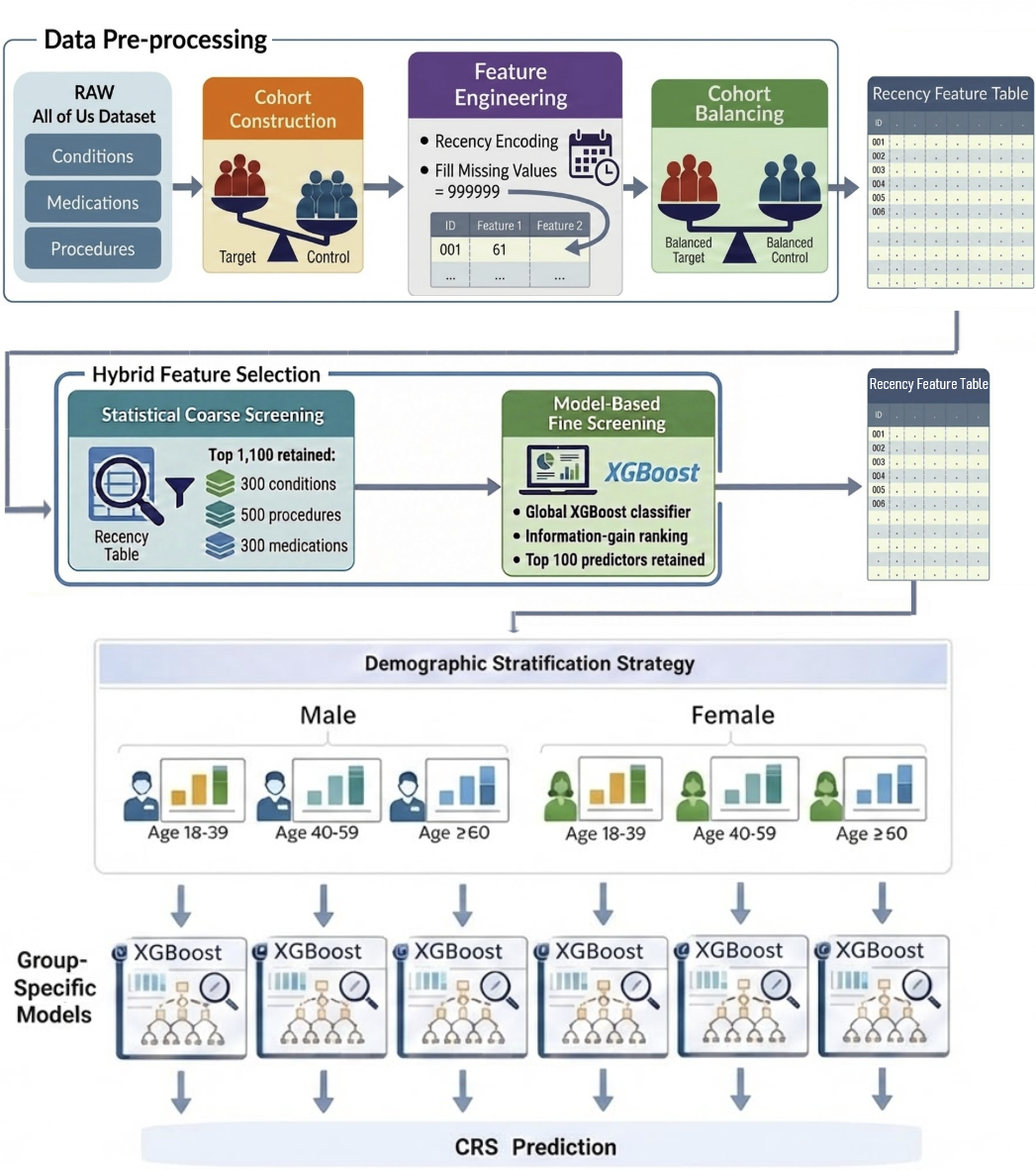}
    \caption{\textbf{Pipeline Overview}. Our framework processes raw data through data pre-processing, identifies key features via hybrid feature selection, and trains group-specific models based on demographic stratification strategies.}
    \label{fig:overall_pipeline}
\vspace{-0.2CM}
\end{figure}

\section{Data Preparation}

To overcome the limited generalizability of prior chronic rhinosinusitis modeling studies which often constrained to single-institution EHRs, we leveraged the NIH \textit{All of Us} Research Program, a nationally diverse, longitudinal cohort with EHR data standardized to the OMOP Common Data Model. However, raw OMOP data are stored as sparse, transactional event logs that are not inherently compatible with standard machine learning algorithms. To bridge this gap, we implemented a rigorous preprocessing pipeline to transform these irregular longitudinal records into structured, recency-based feature representations. This transformation was critical for capturing the temporal dynamics of the disease; by aggregating historical clinical events into fixed-window vectors based on their proximity to the index date, we enabled the model to prioritize recent, active clinical signals over distant history. This process effectively converted raw temporal sequences into a unified recency feature table suitable for predictive modeling.

\vspace{-0.1cm}
\subsection{Cohort Construction and Target Definition}

CRS targets were identified using a rule-based phenotyping strategy. To validate the diagnosis and minimize misclassification from isolated coding events, participants were labeled as CRS targets if they had at least two qualifying CRS-related diagnostic codes recorded within a two-year period. All remaining participants who did not meet the CRS target definition were labeled as controls. 

Controls were propensity score–matched to CRS targets to balance demographics and healthcare utilization, using age category, sex at birth, race, ethnicity, and visit frequency category. Visit frequency was included as a matching criterion to mitigate surveillance (ascertainment) bias, since participants with more healthcare encounters have more opportunities to receive CRS-related diagnoses and recorded comorbidities. Visit frequency was defined by the average number of relevant healthcare visits within a specified timeframe and categorized as fewer than 12 visits, 12–24 visits, or more than 24 visits. For CRS Targets, the index date was defined as the first qualifying CRS diagnosis date; for matched controls, the index date was assigned as the same date as their matched CRS target to ensure aligned observation windows.


\subsection{Cohort Balancing and Feature Engineering}

Raw electronic health records inherently exhibit severe class imbalance, with CRS targets representing a small minority of the general population. Training machine learning models on such skewed data typically results in biased classifiers that prioritize specificity (predicting the majority class) at the expense of sensitivity (detecting true targets). To mitigate this and ensure the model effectively learns the signal of the minority class, we employed a down sampling strategy. 

First, participants without any recorded clinical events during the two-year observation window were excluded to ensure adequate longitudinal information for reliable prediction. Subsequently, from the remaining pool, controls were then matched 1:1 to CRS targets using nearest-neighbor matching based on pre-computed propensity score. This process yielded a final balanced cohort of 17,560 participants ($n_{\text{target}} = 8,780$, $n_{\text{control}} = 8,780$) serving as the foundation for both global model training and subgroup analyses.

To transform raw OMOP data logs into a usable format, we applied the same recency-based encoding scheme across all domains, including conditions, medications, and procedures. Specifically, for diagnostic procedures such as endoscopy and imaging, these features capture the \textit{utilization} of the service (i.e., the occurrence of the event) rather than the clinical results or findings. Unlike traditional binary encoding which merely indicates the presence or absence of a condition, recency encoding captures the temporal proximity of clinical events. This approach mirrors clinical decision-making principles, where recent medical interactions typically hold greater prognostic value and relevance to the patient's current health status than distant historical events.
For each clinical variable, the feature value was defined as the number of days between the most recent occurrence of the event and the index date. To handle sparsity, events that never occurred or occurred outside the two-year observation window were assigned a sentinel value of 999,999. This allows tree-based models to mathematically segregate missing or remote events from recent, clinically active signals without imputing artificial means. Finally, to ensure rigorous evaluation and prevent data leakage, all explicit CRS diagnostic codes were removed from the feature set prior to training.

\begin{figure}[htbp]
    \centering
    \includegraphics[width=0.7\columnwidth]{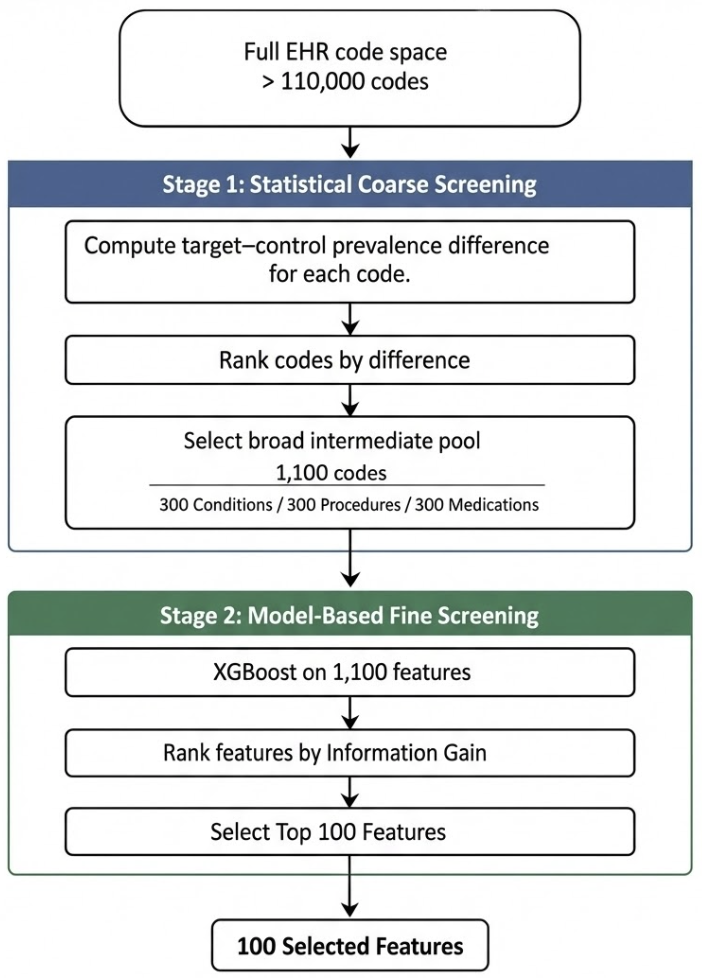}
    \caption{Pipeline of Hybrid Feature Selection}
    \label{fig:Feature Extraction}
\vspace{-0.2CM}
\end{figure}

\section{Methodology}
\label{sec:methodology}

As illustrated in the overall pipeline of our approach (Fig. \ref{fig:overall_pipeline}), the proposed methodology consists of three core components built upon pre-processed data.
First, we build a hybrid feature selection framework to select informative features from extremely high-dimensional data.
Second, we mitigate demographic-dependent heterogeneity by stratifying participants into six sex-age subgroups. 
Third, we develop stratum-specific predictive models using XGBoost, with independent hyperparameter optimization for each training regime, resulting in six independently tuned models aligned with clinically interpretable demographic strata.

\subsection{Hybrid Feature Selection}

In order to improve model performance by reducing noise and dimensionality while preserving clinical interpretability, we adopted a two-stage hybrid feature selection framework that integrates an initial statistical coarse screening with a subsequent model-driven fine screening to progressively narrow the feature space as shown in Fig. \ref{fig:Feature Extraction}.

\textbf{Stage 1: Statistical Coarse Screening}

The raw \textit{All of Us} dataset includes $>$110,000 unique standard codes spanning conditions, procedures, and medications. Modeling such a high-dimensional, sparse feature space is computationally costly and increases overfitting risk. We therefore applied a prevalence-based coarse screen to prioritize variables with the strongest marginal separation between CRS targets and controls.

For each code, we computed prevalence in the target group and control group and used the absolute prevalence difference, $|P_{target}-P_{control}|$, as a simple utility score. Features were ranked by this metric, and the top 1,100 were retained to form an intermediate set: 300 conditions, 500 procedures, and 300 medications. We intentionally kept this intermediate set relatively broad to avoid discarding weaker yet biologically meaningful signals. A larger quota was allocated to procedures to reflect their greater coding granularity and heterogeneity, which can be informative for distinguishing CRS care trajectories. Guided by established epidemiological literature~\cite{sedaghat2022epidemiology, min2015risk} and discussions with clinical experts, we ensured that our feature set comprehensively encompasses the known medical risk factors for CRS.

\textbf{Stage 2: Model-Based Fine Screening}

The prevalence screen captures primarily univariate, near-linear signals. To identify features that contribute under multivariate and potentially non-linear interactions, we performed a second-stage selection using an optimized XGBoost model.

Using the 1,100-feature set, we trained a global XGBoost classifier and ranked features by information-gain-based importance. We retained the top 100 features for downstream modeling to balance parsimony, interpretability, and performance, reducing residual noise while preserving signals relevant to demographic-stratified analyses. Quantitatively, this subset accounts for 80.6\% of the total cumulative SHAP value, ensuring that the vast majority of the model's predictive power is preserved.

The final feature set spanned clinically coherent CRS signals, including sinonasal conditions like: acute sinusitis, chronic rhinitis, allergic rhinitis; diagnostic procedures, such as the utilization of nasal endoscopy, maxillofacial CT; and common CRS medications, such asfluticasone propionate nasal spray, amoxicillin–clavulanate, oral prednisone.

\begin{table}[t]
\centering
\caption{Statistics of the study cohort by demographic group.}
\label{tab:baseline_characteristics}
{%
\begin{tabular}{lcc}
\hline
Sub-group & number & No. of CRS targets \\
\hline
 Male 18-40 & 291 & 174 (59.8\%) \\
 Male 40-60 &  1,184& 598 (50.5\%) \\
 Male 60+ & 3,566 & 1,782 (50.0\%) \\
Female 18-40 & 1,295 & 619 (47.8\%)\\
Female 40-60 &  4,089& 2,039 (49.9\%)\\
Female 60+ & 7,135 & 3,568 (50.0\%)\\
\hline
\end{tabular}%
}
\vspace{-0.2CM}
\end{table}


\subsection{Demographic Stratification Strategy}

CRS is clinically and biologically heterogeneous, and key disease signals can vary systematically across demographic subgroups. Consistent with established epidemiological and biological evidence, our cohort exhibited sex- and age-dependent differences in the dominant families of CRS-related signals.

In our cohort, sex-conditioned enrichment patterns suggested distinct etiologic profiles. Males more often showed a predominantly \emph{structural and polypoid} pattern, with greater prevalence of obstruction- and polyp-associated signals, but females more often exhibited a \emph{mucosal inflammation and atopy} pattern, with greater prevalence of mucosal-inflammatory and allergy/atopy-associated signals. Prior work reporting sex-dependent CRS phenotypes and biology provides external support for these sex-linked signal patterns \cite{stevens2019dysregulation, lu2021sex, wu2018estrogen}.

We further observed age-modulated shifts in dominant signal families conditional on sex. In males, structural/anatomic signals were more prominent in midlife, while polyp-associated signals became more prominent in older age, consistent with reports of male predilection for polyp-predominant CRS and higher procedural burden in CRSwNP populations \cite{tan2013incidence, sowerby2013epidemiology}. In females, acute localized inflammatory or infectious signals were more prominent at younger and middle ages, whereas older females exhibited greater enrichment of chronic multi-airway comorbidity signals. This pattern is supported by evidence of aging-related inflammatory remodeling and bacteria-associated CRS profiles \cite{anderson2022aging, hwang2019chronic}, together with the 'united airway' framework linking upper- and lower-airway disease and bronchiectasis comorbidity \cite{guilemany2009united, maio2018bronchiectasis}.





Based on these observations, the balanced cohort was partitioned into six mutually exclusive strata defined by the intersection of sex and age. Age thresholds of 40 and 60 years were used to define young (age $<40$), middle-aged ($40 \le \text{age} < 60$), and older (age $\ge 60$) subgroups, consistent with commonly adopted categorizations in prior rhinology and chronic disease studies \cite{tan2018age, kim2020age, cho2012age}. Supporting this stratification strategy, a Kruskal-Wallis analysis of our cohort revealed that 87 of the top 100 features exhibited statistically significant distributional differences across these six subgroups ($p < 0.05$), confirming substantial heterogeneity. Six independent predictive models were then trained, one for each demographic stratum. Baseline characteristics of the stratified cohorts are summarized in Table~\ref{tab:baseline_characteristics}.

\subsection{Model Development}

To better capture the feature–outcome relationships and data distributions, we constructed an independent model for each strata, allowing each model to learn tailored decision boundaries that reflect the clinical and temporal characteristics unique to its subgroup.

Given the complex, non-linear interactions among longitudinal clinical features and their heterogeneous effects across patients, we adopted Extreme Gradient Boosting (XGBoost) with decision-tree base learners \cite{chen2016xgboost}. XGBoost is well suited to sparse, high-dimensional EHR data, as it flexibly captures non-linear relationships while remaining robust to heterogeneous feature scales. A key advantage in our setting is its built-in sparsity-aware split finding mechanism, which enables principled handling of sentinel-valued features without ad hoc imputation. During tree construction, sentinel values are treated as a distinct category and are automatically routed to the split direction that maximizes information gain. This property is particularly important for our recency-encoded features, where a sentinel value ($999{,}999$) represents the absence of prior events, allowing the model to distinguish patients with no history from those with distant or recent events while controlling model complexity through regularization.

We formulated CRS status prediction as a supervised binary classification task and modeled it using XGBoost. At boosting iteration $t$, the algorithm learns an additive tree ensemble by minimizing a regularized objective:
\begin{equation}
\mathcal{L}^{(t)}=\sum_{i=1}^{n} l\left(y_i,\hat{y}^{(t-1)}_i+f_t(x_i)\right)+\Omega(f_t),
\end{equation}
where $l(\cdot)$ denotes the loss function and $\Omega(\cdot)$ penalizes model complexity to encourage parsimonious trees. Each base learner $f_t$ is a regression tree fitted to the first- and second-order gradients of the loss, enabling efficient optimization and stable convergence. The regularization term constrains tree structure and leaf weights, allowing the model to capture non-linear feature interactions while reducing overfitting in high-dimensional longitudinal clinical data.

\section{Experiment}
\subsection{Experiment settings}

All sex-by-age–stratified models were trained on a single-machine CPU environment with multi-threaded parallelization. For each demographic subgroup, an independent XGBoost model was developed, with hyperparameters optimized using the Bayesian Tree-structured Parzen Estimator (TPE) algorithm over 100 trials to maximize the Area Under the Receiver Operating Characteristic curve (AUC-ROC). This subgroup-specific optimization enabled both hyperparameter values and structural constraints to adapt to variations in risk factor distributions and data density across subpopulations. The final optimized hyperparameter configurations for these tailored models are summarized in Table \ref{tab:group_hyperparameters}. Model performance and hyperparameter robustness were assessed using 10-fold stratified cross-validation.

\begin{table}[t]
\centering
\caption{Hyperparameter of demographic-specific XGBoost models.}
\label{tab:group_hyperparameters}
\resizebox{\columnwidth}{!}{%
\begin{tabular}{lcccccc}
\hline
Group & n\_est & depth & learning rate & subsamp & colsamp & reg\_$\alpha$ \\
\hline
Female 18-40 & 212 & 10 & 0.020 & 0.97 & 0.61 & 0.3400  \\
Female 40-60 & 280 & 7  & 0.038 & 0.65 & 0.70 & 0.0005 \\
Female 60+   & 290 & 8  & 0.032 & 0.62 & 0.68 & 0.0002 \\
Male 18-40   & 160 & 6  & 0.022 & 0.93 & 0.61 & 0.0610  \\
Male 40-60   & 200 & 7  & 0.028 & 0.78 & 0.72 & 0.0250  \\
Male 60+     & 265 & 8  & 0.036 & 0.58 & 0.62 & 0.0003 \\
\hline
\end{tabular}%
}
\vspace{-0.2CM}
\end{table}

Our proposed method was systematically compared with Random Forest, Support Vector Machine (SVM), and deep learning models, all trained and evaluated on the same dataset to ensure a fair and consistent comparison. Model performance was assessed using a comprehensive suite of evaluation metrics from multiple perspectives: (1) AUC-ROC to quantify the overall discriminative ability of the model; (2) F1-score to reflect performance under class imbalance; (3) Sensitivity (Recall) to measure the ability to correctly identify targets; and (4) Specificity to evaluate the capability to correctly recognize controls.
For binary classification, a threshold of 0.5 was applied to generate the confusion matrices.


\subsection{Results}
Among the evaluated architectures, our proposed model demonstrated the strongest overall discriminative ability, achieving AUC of \textbf{0.8461} (Table \ref{tab:overall_performance}) and outperforming the comparative baselines, including XGBoost (0.8293), Random Forest (0.8159), SVM (0.8120), and Deep Learning (0.8072). This performance superiority extends to the F1-score, where our model reached \textbf{0.7415}, indicating a more robust balance between precision and recall compared to the next best performer (XGBoost, 0.7208). Crucially, our method achieved the highest sensitivity of \textbf{0.6889}, significantly surpassing the standard machine learning approaches. Although specificity was lower than the global baseline, this trade-off is clinically advantageous, as higher sensitivity is prioritized in diagnostic screening to ensure affected patients are not overlooked.

The comparative analysis reveals that our demographic-specific XGBoost models consistently outperformed the global model across every evaluated subgroup (Table \ref{tab:group_performance}). Specifically, the tailored models achieved higher AUC values in all six strata, with improvements ranging from $+0.0076$ to $+0.0278$. This demonstrates that a stratified modeling strategy preserves distinct risk signals that are otherwise diluted in a unified global approach.

\begin{table}[t]
\centering
\caption{Overall performance comparison}
\label{tab:overall_performance}
\begin{tabular}{lcccc}
\toprule
\textbf{Method} & \textbf{AUC} & \textbf{F1} & \textbf{Sensitivity}& \textbf{Specificity} \\ 
\midrule
Random Forest & 0.8159 & 0.7073 & 0.6378 & 0.8343 \\
SVM & 0.8120 & 0.7082 & 0.6442 & 0.8249 \\
Deep Learning & 0.8072 & 0.7124 & 0.6724 & 0.7846 \\
XGBoost & 0.8293 & 0.7208 & 0.6464 & 0.8531 \\
\midrule
Ours & \textbf{0.8461} & \textbf{0.7415} & \textbf{0.6889} & 0.8240 \\
\bottomrule
\end{tabular}
\vspace{-0.1CM}
\end{table}

\begin{table}[t]
\centering
\caption{Performance comparison across subgroups between the global XGBoost model and demographic-specific XGBoost models.}
\label{tab:group_performance}
\begin{tabular}{lcccc}
\toprule
\textbf{Group} & \textbf{\% of Cohort} & \textbf{AUC(Glo)} & \textbf{AUC(Gro)}& \textbf{$\Delta$ AUC} \\ 
\midrule
Female 18--40 & 7.4\%  & 0.8713 & 0.8789 & +0.0076 \\
Female 40--60 & 23.3\% & 0.7940 & 0.8218 & +0.0278 \\
Female 60+    & 40.6\% & 0.8410 & 0.8546 & +0.0136 \\
Male 18--40   & 1.7\%  & 0.9286 & 0.9548 & +0.0262 \\
Male 40--60   & 6.7\%  & 0.8109 & 0.8225 & +0.0116 \\
Male 60+      & 20.3\% & 0.8291 & 0.8435 & +0.0144 \\ 
\midrule
\textbf{Total} & \textbf{100\%} & \textbf{0.8293} & \textbf{0.8461} & \textbf{+0.0168} \\
\bottomrule
\end{tabular}
\vspace{-0.2CM}
\end{table}

\section{Discussion}

Our analysis of the \textit{All of Us} cohort emphasizes that CRS manifests as a spectrum of clinical phenotypes modulated by biological sex and age. Stratifying this cohort highlights divergences in clinical profiles, ranging from anatomical obstruction in males to systemic airway inflammation in older females. These findings support the utility of the stratified modeling approach described in Section~\ref{sec:methodology} and appear consistent with epidemiological precedents regarding the heterogeneity of sinonasal disease \cite{chiu2026biological}.

\subsection{Clinical Phenotypes and Demographics}

Distinct demographic trajectories were observed regarding potential disease etiology, based on prevalence comparisons between each subgroup and the remainder of the cohort. The male cohort demonstrated an age-dependent divergence characterized largely by structural and polypoid pathology. Middle-aged males (40-60) exhibited a predominantly anatomical profile, showing a significantly elevated rate of septal deviation as compared to the other demographic sub-groups (4.73\% vs 2.20\%; $P < 0.001$). In contrast, elderly males ($\ge$60) exhibited a significantly higher prevalence of nasal polyps (1.60\% vs 0.82\%; $P < 0.001$), observations that align with the established predilection for CRSwNP and may reflect a need for intensive surgical management \cite{tan2013incidence, sowerby2013epidemiology}.

Conversely, the female cohort exhibited profiles characterized more by mucosal inflammation and atopy, with age modulating the transition from acute to systemic presentation. Young females (18-40) exhibited elevated rates of otitis media (5.56\% vs 2.72\%; $P < 0.001$), suggesting a phenotype po-tentially linked to Eustachian tube dysfunction. Middle-aged females (40-60) constituted the primary demographic for acute inflammatory events, with a significantly higher incidence of acute frontal sinusitis (1.69\% vs 0.83\%; $P < 0.001$). The elderly female stratum ($\ge$60) shifted toward chronic inflammation and an atopic picture, exhibiting elevated rates of bronchiectasis (1.33\% vs 0.86\%; $P = 0.004$) and fexofenadine usage (1.04\% vs 0.70\%; $P = 0.020$). These observations parallel the concept of a systemic inflammatory burden affecting both the upper and lower airways in older populations \cite{guilemany2009united, maio2018bronchiectasis}. 

Collectively, these distinct phenotypic profiles indicate possible areas for early management and intervention strategies based on sex; such as addressing structural abnormalities in males and optimizing allergic or inflammatory symptom management in females with suspected CRS.

\subsection{Phenotypic Consistency and Model Adaptation}

The feature importance rankings derived from the stratified models mirror observed demographic divergences, implying the capture of biologically relevant signals. Male-specific models prioritized anatomical features: \textit{deviated nasal septum} was a dominant predictor (Rank 2) for younger men (18--40), shifting to \textit{polyp of nasal cavity} (Rank 3) in elderly males ($\ge$60). In contrast, female models favored mucosal and inflammatory indicators. The Female 40--60 model prioritized \textit{allergic rhinitis} (Rank 5) and \textit{asthma} (Rank 8), while the Female 60+ group emphasized infectious markers, reinforcing a distinct inflammatory etiology over mechanical obstruction.

Beyond feature selection, hyperparameter optimization illustrated the algorithm's adaptation to data availability. In smaller subgroups, more complex tree ensembles more readily fit stochastic fluctuations and sparse co-occurrence patterns, so tuning favored conservative capacity control (e.g., shallower depth, stronger regularization, larger minimum leaf size, and stronger subsampling). In larger subgroups, the data supported higher-capacity configurations to capture nonlinearities and interactions while maintaining stable generalization.

\section{Conclusion}

In this work, we established a predictive framework for Chronic Rhinosinusitis (CRS) using longitudinal electronic health record (EHR) data from the nationwide \textit{All of Us} Research Program. We transformed sparse OMOP event logs into recency-encoded features and used a hybrid selection pipeline to reduce 110,000 candidate codes to 100 interpretable predictors. To reflect population heterogeneity, we trained sex-by-age--stratified XGBoost models with subgroup-specific hyperparameter tuning. The grouped strategy improved weighted AUC by 0.0168 to 0.8461, with gains in every stratum. Overall, the results of the work suggest that routine EHR data may enable actionable pre-diagnostic risk stratification among primary care physicians to prioritize referral and early interventions.



\newpage
\bibliographystyle{plain}  
\bibliography{references}  

\end{document}